\documentclass[letterpaper, 10 pt, conference]{ieeeconf}  

\newcommand{\approxtilde}{\raisebox{0.5ex}{\texttildelow}}
\newcommand\marklessfootnote[1]{
    \addtocounter{footnote}{1} 
    \footnotetext{#1}
}

\usepackage{graphicx}
\usepackage{amsfonts}
\usepackage{amsmath}
\usepackage{balance}
\makeatletter
\let\NAT@parse\undefined
\makeatother
\usepackage{cite}
\usepackage[colorlinks,allcolors=magenta]{hyperref} 
\usepackage[all]{hypcap}
\newcommand{\smalltt}[1]{{\small \ttfamily #1}}
\usepackage{tikz}
\newcommand\copyrighttext{%
  \footnotesize \textcopyright \the\year{} IEEE. Personal use of this material is permitted. Permission from IEEE must be obtained for all other uses, in any current or future media, including reprinting/republishing this material for advertising or promotional purposes, creating new collective works, for resale or redistribution to servers or lists, or reuse of any copyrighted component of this work in other works.}
\newcommand\copyrightnotice{%
\begin{tikzpicture}[remember picture,overlay]
\node[anchor=south,yshift=10pt, text width=0.85\textwidth, align=justify, draw] at (current page.south) {\copyrighttext}; 
\end{tikzpicture}%
\vspace{-10.25pt}
}

\IEEEoverridecommandlockouts                              

\title{\LARGE \bf
The Monado SLAM Dataset for Egocentric Visual-Inertial Tracking
}

\author{Mateo de Mayo$^{1}$$^,$$^{2}$, Daniel Cremers$^{1}$, Taihú Pire$^{3}$
\thanks{$^{1}$Mateo de Mayo and Daniel Cremers are with the Technical University of Munich, and the Munich Center for Machine Learning, Munich, Germany
        {\tt \{mateo.demayo, cremers\}@tum.de}}
\thanks{$^{2}$Mateo de Mayo was previously with Collabora Ltd. Cambridge, UK}
\thanks{$^{3}$Taihú Pire is with the CIFASIS, CONICET-UNR, Rosario, Argentina
        {\tt pire@cifasis-concicet.gov.ar}}%
}

\begin{document}

\maketitle
\copyrightnotice
\thispagestyle{empty}
\pagestyle{empty}

\begin{abstract}

Humanoid robots and mixed reality headsets benefit from the use of head-mounted
sensors for tracking. While advancements in visual-inertial odometry (VIO) and
simultaneous localization and mapping (SLAM) have produced new and high-quality
state-of-the-art tracking systems, we show that these are still unable to
gracefully handle many of the challenging settings presented in the head-mounted
use cases. Common scenarios like high-intensity motions, dynamic occlusions,
long tracking sessions, low-textured areas, adverse lighting conditions,
saturation of sensors, to name a few, continue to be covered poorly by existing
datasets in the literature. In this way, systems may inadvertently overlook
these essential real-world issues. To address this, we present the Monado SLAM
dataset, a set of real sequences taken from multiple virtual reality headsets.
We release the dataset under a permissive CC BY 4.0 license, to drive
advancements in VIO/SLAM research and development.

\end{abstract}


\section{INTRODUCTION}

\setcounter{footnote}{3}

Egocentric visual-inertial tracking methods, such as visual-inertial odometry
and SLAM (VIO/VI-SLAM) are essential for applications in humanoid robotics, and
augmented and virtual reality (AR/VR), known collectively as XR. Evaluating
these methods typically involves the use of datasets collected from a
specialized sensor rig. Samples from cameras and inertial measurement units
(IMU) are recorded alongside a ground-truth trajectory of the rig. This
ground-truth can then be used to measure the accuracy of a trajectory estimate
from a given system. Different kinds of datasets are available, featuring hand-held
devices \cite{zhangHiltiOxfordDatasetMillimeterAccurate2023}, car-mounted
cameras \cite{geigerVisionMeetsRobotics2013a}, and UAVs
\cite{burriEuRoCMicroAerial2016}. Head-mounted XR datasets can also be found, but
recordings from only two AR devices exist in the literature
\cite{ungureanuHoloLens2Research2020,engelProjectAriaNew2023}. While
these works have their own merits, they can prove insufficient for stressing the
accuracy of current state-of-the-art systems. Furthermore, none of these
datasets cover the head-mounted VR use case, which can be particularly
challenging, due to the ability to provide immersive and high-intensity
simulations to the operator.

\begin{figure}
  \includegraphics[height=0.525 \textwidth]{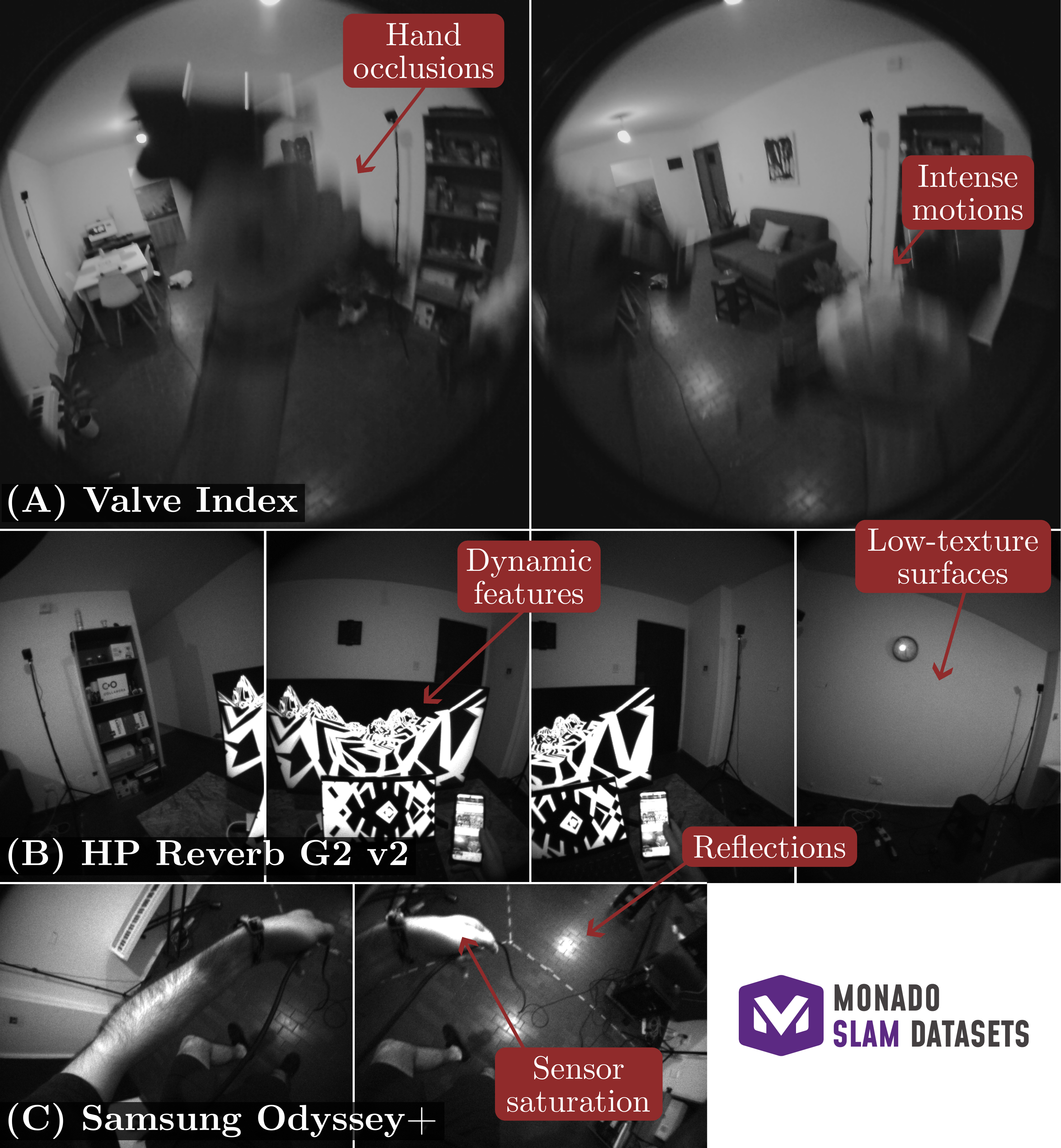}
  \caption{Example views from the Monado SLAM dataset. Each row corresponds to images from
  a distinct device. Image sizes are proportional to their resolution. We
  highlight in red some examples of the challenges present
  in the dataset.}
  \label{fig:cameras}
\end{figure}

To address these limitations and help further advancements in the VIO/VI-SLAM
field, we present the \textbf{Monado SLAM dataset (MSD)}. MSD presents a challenging and
diverse set of sequences on a standard VR setup. It is recorded using three
different commercial headsets for increased data
diversity as can be seen in Fig. \ref{fig:cameras}: a Valve Index, an HP Reverb G2, and a Samsung Odyssey+.
All datasets are pre-calibrated, and their coordinate frames and timestamps are
aligned by default.
The ground-truth is captured on a setup with 3 Lighthouse base stations which
provide an average accuracy of \approxtilde1 cm
\cite{holzwarthComparingAccuracyPrecision2021,borgesHTCViveAnalysis2018}.
We leverage the Monado open-source
platform\footnote{\label{foot:monado}\url{https://monado.freedesktop.org}} to capture sensor data
from these XR devices. A preliminary version of this dataset has already been
used in \cite{behrooziSlimSLAMAdaptiveRuntime2024}, showing its applicability in
the field.

In this release, we present 64 non-calibration recordings with 5 hours and 15
minutes of real-world footage and challenging scenarios, including long and
uninterrupted datasets of up to \approxtilde40 minutes and fast-paced gameplay.
Additionally, we contribute a thorough benchmark, establishing a
baseline for future research. We evaluate the dataset on a diverse set of
top-scoring VIO/SLAM systems covering different techniques and modalities.
Although these systems are very capable, MSD is able to highlight their
weaknesses for live operation in multiple common scenarios.

\begin{table*}[h!]
  \caption{Overview of visual-inertial datasets}
  \includegraphics[width=\textwidth]{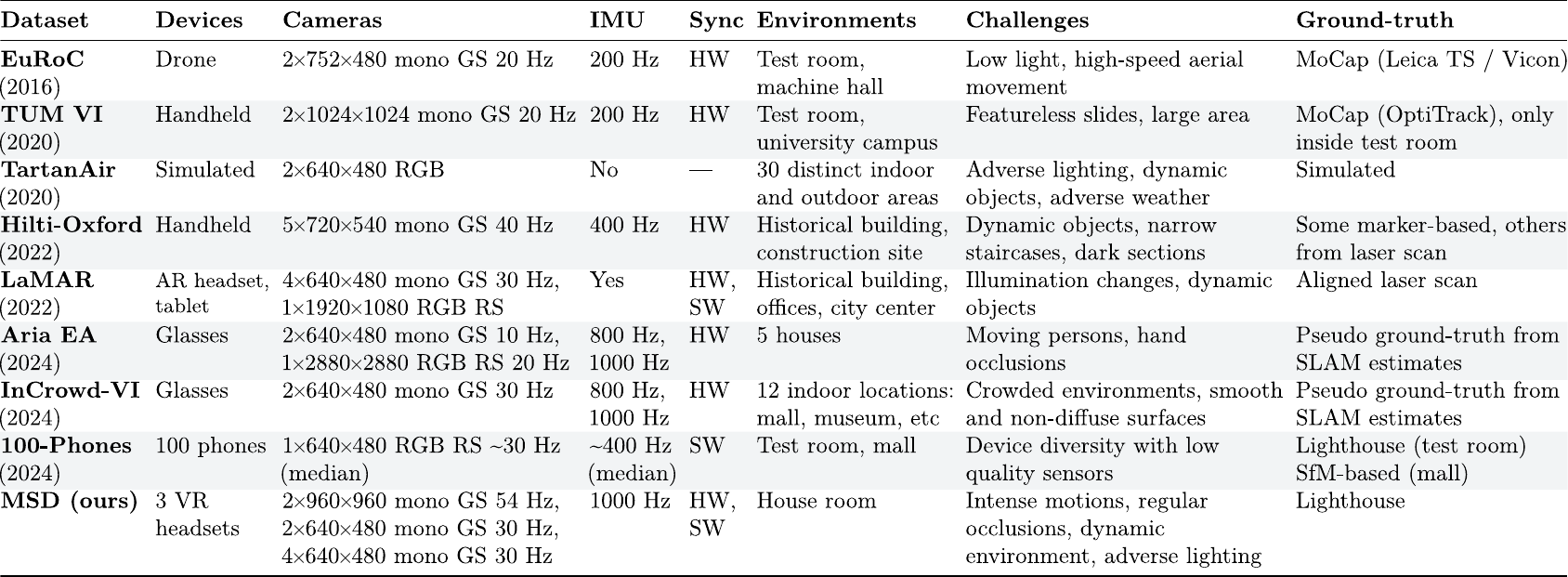}
  \centering
  \textit{GS: global shutter \;\; RS: rolling shutter \;\; mono: monochrome \;\; HW: hardware time synchronization \;\; SW: software time synchronization.}
  \label{fig:datasettable}
\end{table*}

The main contributions of this work are the following:

\begin{itemize}
  \item A challenging and permissively licensed visual-inertial dataset that
  reveals limitations of current systems.
  \item The first VI dataset recorded with VR devices, allowing highly
  dynamic motions through intense gameplay.
  \item Extending the scarce available VI data for XR tracking of only two XR
  devices with three distinct headsets.
  \item A cost-efficient methodology for high-quality dataset generation. We
  make all our tools
  available\footnote{\label{foot:xrtslam}\url{https://gitlab.freedesktop.org/mateosss/xrtslam-metrics}}.
  \item A thorough benchmark on state-of-the-art systems, with a
  focus on metrics relevant to online (causal) operation.
\end{itemize}

We release MSD under the permissive CC BY 4.0 license, to foster collaborations
between scientific communities and industry. The full dataset, including raw and
calibrated data, and its documentation is available on:
\begin{center}
  \texttt{\url{https://huggingface.co/datasets/collabora/monado-slam-datasets}}
\end{center}%

\section{RELATED WORK}

Visual-inertial datasets have paved the way for the development of more accurate
and robust tracking systems. The challenges covered by these datasets influence
the focus of new research systems since the data provides a
benchmark upon which to improve. In the context of VI tracking,
collecting high-quality footage is not trivial due to the specific data
requirements. Good recording platforms require global shutter cameras,
hardware time synchronization, rigid build materials, and accurate ground-truth
collection. Coming up with new platforms that satisfy these
specialized requirements can prove to be a complex task.

For XR, this difficulty is even more pronounced since head-mounted devices
require additional levels of design to also accommodate high-resolution
displays, lenses, audio systems, and comfortable interfaces for users to wear.
Commercial headsets that have taken care of the design for such constraints do
exist, but for most, there is no standard way to access and record their raw
sensor data. To the best of our knowledge, publicly available VI datasets that
cover head-mounted XR use-cases only exist for two devices: the \textit{HoloLens
2} AR headset
\cite{chandioHoloSetDatasetVisualInertial2023,sarlinLaMARBenchmarkingLocalization2022},
and the \textit{Project Aria glasses} \cite{engelProjectAriaNew2023}. This means, the Monado
SLAM dataset is the first dataset containing
recordings from \textit{VR} devices, with three distinct headsets, which account
for much more challenging types of sequences due to the use of immersive and
high-intensity VR simulations.

In this section, we describe different visual-inertial datasets from the
literature. Since no other VR dataset exists, and only few are head-mounted,
we also consider more general works that have made significant contributions
to the VI tracking field. Table \ref{fig:datasettable} shows an overview of notable
characteristics for each of these publications.

The \textbf{EuRoC} ASL dataset \cite{burriEuRoCMicroAerial2016} is one of the most widespread benchmarks used in the
field. It presents UAV navigation captures with a reasonable degree of
difficulty. While not as challenging as sequences present in MSD, to
this day, it still is a reference benchmark that can showcase inaccuracies in
tracking systems. It comprises 11 characteristic sequences with a total of \approxtilde20
minutes of stereo-inertial data.

The \textbf{TUM-VI} dataset \cite{schubertBasaltTUMVI2018} is another highly influential work in the area. It presents
\approxtilde1.5 hours of hand-held recordings in different environments. It is also one of the few
benchmarks with photometric calibration available, something particularly
important for direct and semi-direct methods. While it covers multiple
environments, its ground-truth is only present in one of the areas, rendering
most of the dataset unfit for studying trajectory estimates in detail. MSD
offers dense ground-truth data throughout the entire duration of its more than 5
hours of footage.

\textbf{TartanAir} \cite{wangTartanAirDatasetPush2020a} is a very large
synthetic dataset generated in Unreal
Engine\footnote{\url{https://www.unrealengine.com}} with more than 4 TB of data.
It provides multiple challenging scenarios, including adverse weather
conditions, moving objects, and aggressive turns. One of the strong points of
the dataset is its coverage of scenarios that are difficult to replicate. Since
it is a simulation, it can fail to capture the nuances and hidden variables of
real world data.

The \textbf{Hilti-Oxford} \cite{zhangHiltiOxfordDatasetMillimeterAccurate2023} dataset presents a hand-held device with five cameras. It
pushed for the adoption of multi-camera support on newer systems. Its
cm-accurate ground-truth is optimized by aligning LiDAR measurements with high-quality
3D scans. It contains challenging scenarios like low illumination and
dynamic objects. It hosts a live leaderboard\footnote{\url{https://hilti-challenge.com/leader-board-2023.html}} that serves
as an important benchmark of current research trends. Unfortunately, its license
disallows commercial usage, rendering its usage in industrial research
difficult.

The HoloLens 2 is one of the few head-mounted devices that have a “research mode”
\cite{ungureanuHoloLens2Research2020} and can readily be used for data collection. \textbf{LaMAR} \cite{sarlinLaMARBenchmarkingLocalization2022} is
a significant AR dataset using this device. Similarly to the Hilti-Oxford dataset, it
provides cm-accurate ground-truth from ICP LiDAR-Scanner alignment. One of the
unaddressed points of this work is the recording of highly dynamic user
movements, since it primarily contains walking trajectories. Furthermore, the
lack of a benchmark obfuscates understanding the performance of current
systems on the dataset. Another dataset captured with the HoloLens 2 is
\textbf{HoloSet} \cite{chandioHoloSetDatasetVisualInertial2023},
but its ground-truth comes from the black-box tracking system provided
by its proprietary platform with unknown precision.

Another notable device in this area is the Project Aria glasses \cite{engelProjectAriaNew2023}, since its main
purpose is to serve as a data collection platform for XR research. Multiple
datasets have been released under the scope of Project Aria
\cite{panAriaDigitalTwin2023b,maNymeriaMassiveCollection2024b,banerjeeHOT3DHandObject2024}, like the
\textbf{Aria Everyday Activities} (EA) dataset \cite{lvAriaEverydayActivities2024a}. These target XR-tasks like
semantic understanding, hand-tracking, and full-body tracking, to name a few.
Unfortunately, they have sequences that are mostly focused on lightweight AR
use-cases. They are usually well illuminated and without many abrupt motions.
Its license also prohibits commercial use. \textbf{InCrowd-VI} \cite{bamdadInCrowdVIRealisticVisual2024a} is another
dataset made with these glasses, focusing on navigating crowded environments.
Its reference trajectories are a pseudo ground-truth obtained from a custom SLAM
service with unknown accuracy.

\textbf{100-Phones} \cite{zhang100PhonesLargeVISLAM2024} is one of the few works focusing on device variety. It provides simple
datasets for 100 different Android phones featuring common use-cases for
hand-held AR. Some of their datasets have dense ground-truth, and similar to
MSD, provide it through the \textit{SteamVR 2.0} platform with external Lighthouse
trackers. \textbf{ADVIO} \cite{cortesADVIOAuthenticDataset2018a} and \textbf{MARViN} \cite{liuMARViNMobileAR2024} are other two notable phone AR datasets.
Unfortunately both provide low camera frequencies at 5 Hz and 1-2 Hz
respectively\footnote{Not all ADVIO's cameras have low frequencies.}. They also
have inadequate ground-truth: with an accuracy between 10 cm and 1 m in ADVIO, and
MARViN having pseudo ground-truth from COLMAP \cite{schonbergerStructurefromMotionRevisited2016} with unspecified accuracy.

These works have helped advance the field in different ways but have their
limitations when looked through the lens of fast-paced and challenging
head-mounted tracking. The Monado SLAM Datasets provide extensive sequences from
multiple commercial devices with dense ground-truth. We cover various
challenging scenarios like high-intensity movements, adverse illumination
conditions, dynamic objects, and frequent occlusions. With all that is provided,
MSD highlights issues in the current state of the art that go beyond the
specifics of XR and generalize to other use-cases. Therefore, this work fills an
important gap present in the current research literature.

\section{DATASET}

MSD footage is captured from three different devices in a single indoor
environment. The sequences cover many scenarios that are to be expected in
head-mounted operation. In this section, we first describe the devices utilized
for recording, as well as the type of data and sequences the dataset contains.
Lastly, we detail the post-processing and calibration procedures to obtain
intrinsics, extrinsics, and timestamp offsets among the sensors.

\subsection{Devices}

\begin{figure}
  \centering
  \includegraphics[width=0.475 \textwidth]{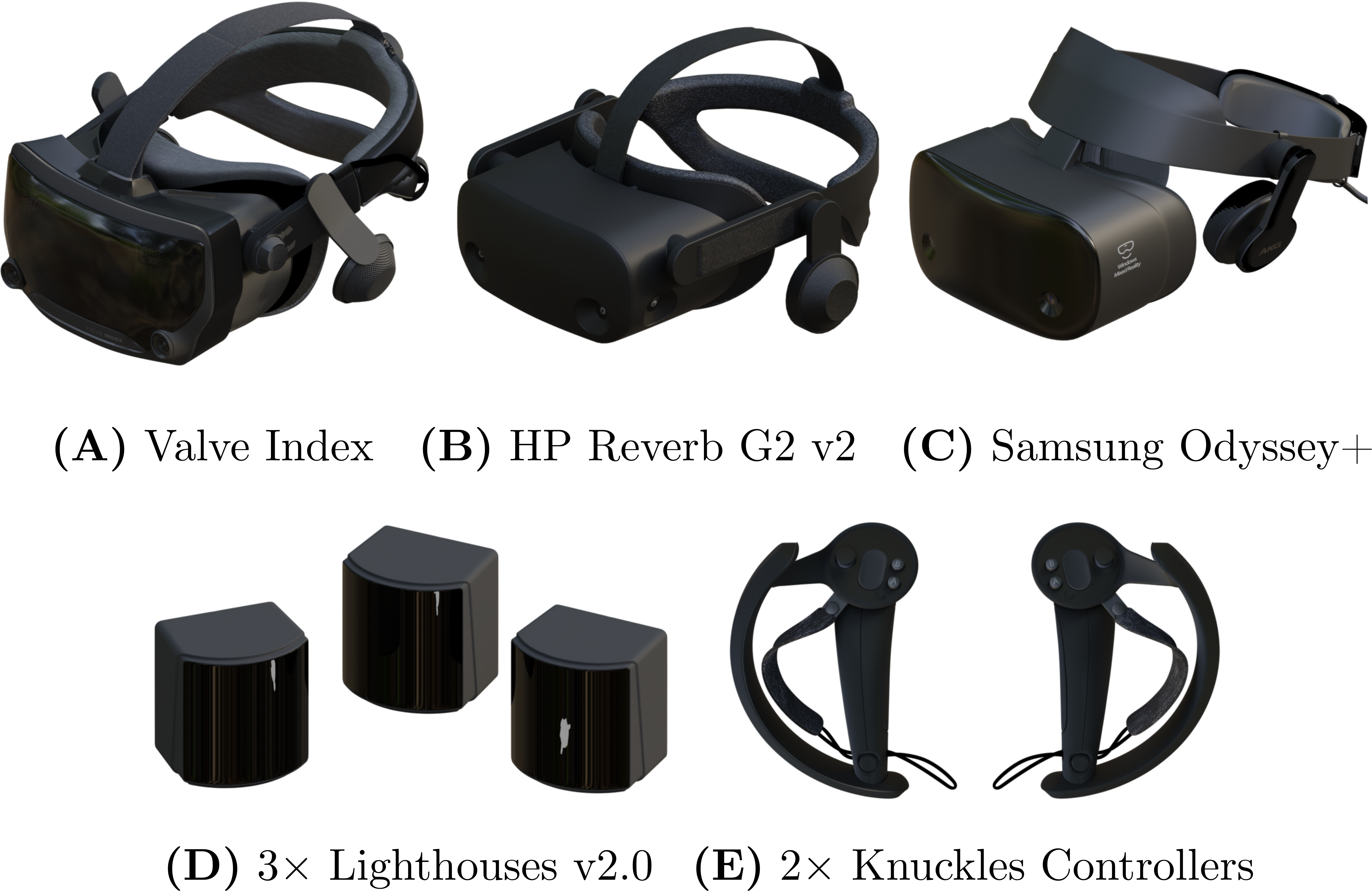}
  \caption{The devices involved in the recording. Three
  different headsets (A, B, C) are used for data collection. Ground-truth
  is obtained from three infrared tracking stations (D). The
  controllers (E) can be tracked by the base stations and are used for gameplay
  input (in A) and for ground-truth collection (in B and C).}
  \label{fig:devices}
\end{figure}

Three consumer headsets were used for the data recording: a \textit{Valve
Index}, an \textit{HP Reverb G2}, and a \textit{Samsung Odyssey+}. They can be
seen in Fig. \ref{fig:devices}. Table \ref{tbl:sensors} shows an overview of the
sensor characteristics of each device. The data shares the same format as the
widespread EuRoC ASL dataset \cite{burriEuRoCMicroAerial2016}. We also make
available tools\textsuperscript{\ref{foot:xrtslam}} to convert
each sequence to rosbag files \cite{macenskiRobotOperatingSystem2022}. The
sequences contain the arrival timestamps of the USB packets to help future
systems simulate real-time operation.
All the footage is recorded through \textit{Monado}\textsuperscript{\ref{foot:monado}}. Monado is an open-source software
platform for XR devices implementing the OpenXR specification
\cite{thekhronosgroupinc.OpenXRSpecification}. Furthermore, it provides the
functionality to operate the headsets, and we contributed to the project
extending it whenever necessary to extract and record the appropriate data.

\marklessfootnote{\label{foot:yuyv422}\url{https://docs.kernel.org/userspace-api/media/v4l/pixfmt-packed-yuv.html}}
\begin{table}
  \caption{Headsets sensor list}
  \includegraphics[width=0.475 \textwidth]{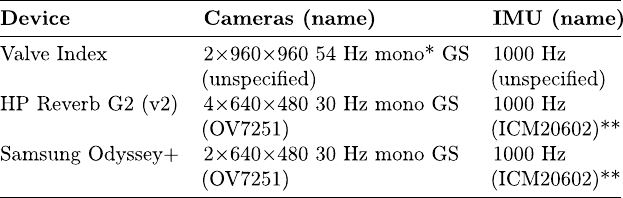}
  \textit{*We store only the “luma” component from a YUYV422\textsuperscript{\ref{foot:yuyv422}} colored image.}
  \newline
  \textit{**Magnetometer available and recorded.}
  \label{tbl:sensors}
\end{table}

\begin{figure}[b]
  \centering
  \includegraphics[width=0.475 \textwidth]{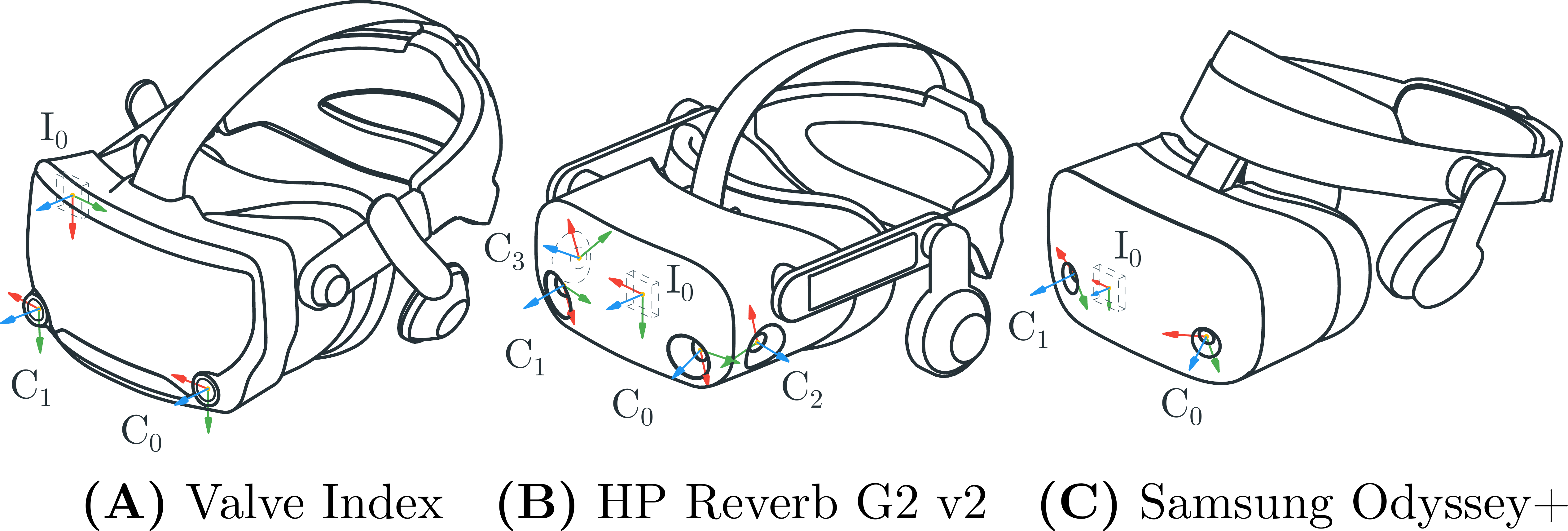}
  \caption{Coordinate systems of the calibrated sensors. In all cases, +Z (blue)
  is camera forward, while +X (red) and +Y (green) follow the right and down
  coordinates in the image pixel grid, respectively. Cameras are named as $C_i$
  ($i=0,1,2,3$) in the diagram and each IMU as $I_0$.}
  \label{fig:schematics}
\end{figure}

The first headset is the \textbf{Valve Index} (Fig. \ref{fig:devices}\,A). It
has an IMU and two forward facing cameras as Fig. \ref{fig:schematics}\,A
shows. Images from this camera can be
seen in Fig. \ref{fig:cameras}\,A. Given their high
resolution and frequency, it is a challenging device to achieve real-time
operation on. This headset and the two
controllers shown in Fig. \ref{fig:devices}\,E are tracked by the SteamVR 2.0
Lighthouse tracking system
\cite{holzwarthComparingAccuracyPrecision2021,taffanelLighthousePositioningSystem2021,borgesHTCViveAnalysis2018}
through three external base stations as seen in Fig. \ref{fig:devices}\,D.

The \textbf{Samsung Odyssey+} (Fig. \ref{fig:devices}\,C) is built on top of the
Windows Mixed Reality
platform\footnote{\label{foot:wmr}\url{https://learn.microsoft.com/en-us/windows/mixed-reality/}}
(WMR). Given its low resolution and frequency, it challenges systems that rely
on high-quality samples. The distribution of its sensors can be seen in Fig.
\ref{fig:schematics}\,C. Only half of their views overlap at common depths (Fig. \ref{fig:cameras}\,C) while
the non-overlapping regions require tracking without stereo matches. This
arrangement turns out to not be supported by many systems, as we will discuss in
section \ref{sec:evaluations}. Magnetometers are a common complementary sensor in
off-the-shelf IMUs, but most state-of-the-art tracking systems do not take advantage
of them even though they can improve tracking accuracy
\cite{joshiEnhancingVisualInertial2024}. MSD includes magnetometer measurements
at 50 Hz for this headset to help future works explore their potential.

The \textbf{HP Reverb G2} (Fig. \ref{fig:devices}\,B) also leverages the WMR platform and shares
the same sensors. This headset, however, has four cameras instead of two,
providing a valuable dataset for multi-camera fusion. While this modality has
been gaining traction \cite{zhangHiltiOxfordDatasetMillimeterAccurate2023},
there are still few systems taking advantage of it
\cite{leuteneggerOKVIS2RealtimeScalable2022,eckenhoffMIMCVINSVersatileResilient2021,usenkoBasaltVisualInertialMapping2020}.
The sensor placement shown in Fig. \ref{fig:schematics}\,B, arranges cameras in
portrait orientation to improve the vertical field of view. The front-facing
cameras have two thirds of their views overlapping at common depths as shown
in Fig. \ref{fig:cameras}\,B. The respective front-and-side camera pairs share
only around a quarter of their views.

\subsection{Sequences}

\begin{figure*}
  \centering
  \includegraphics[width=\textwidth]{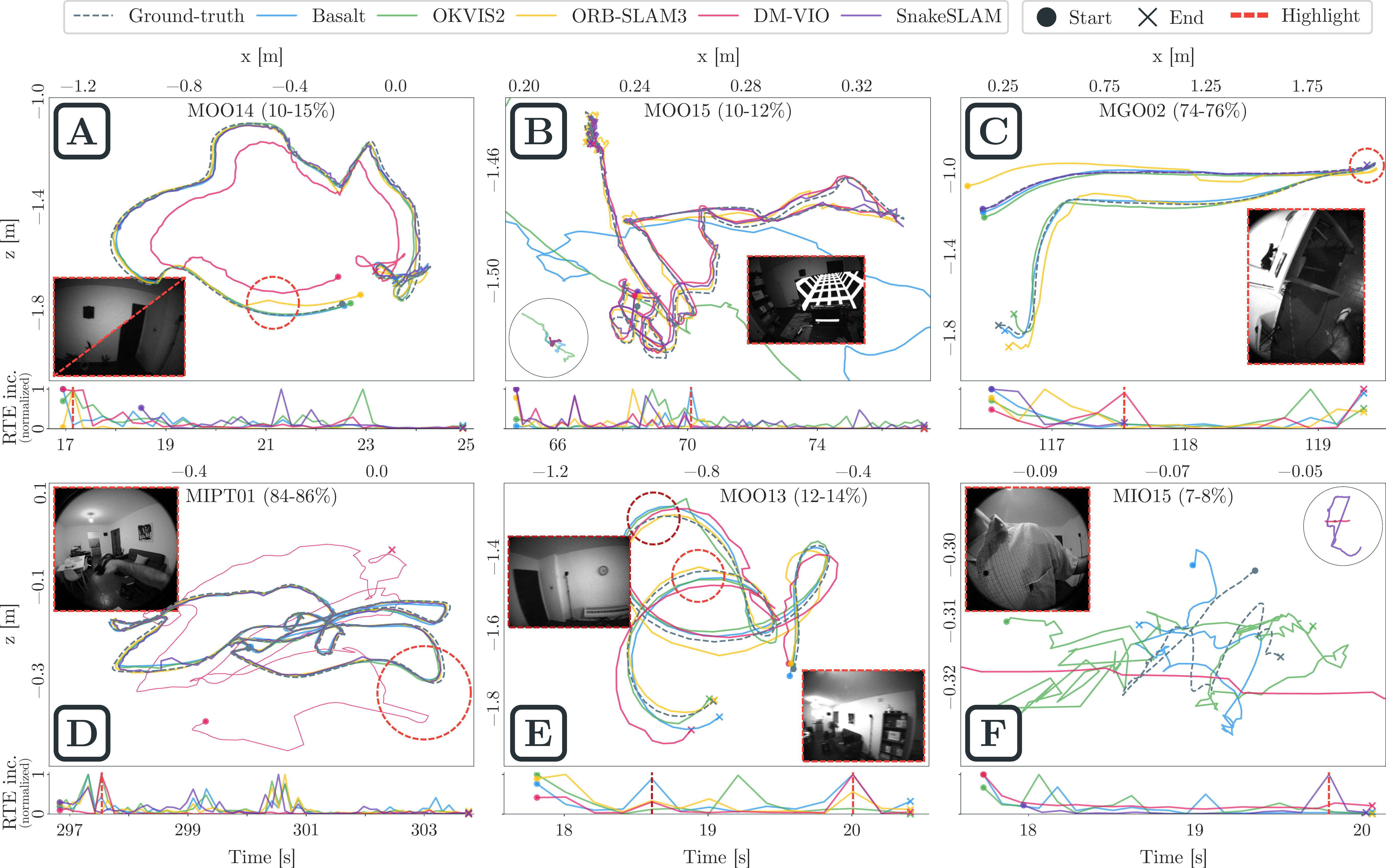}
  \caption{Example subtrajectories of the dataset with normalized RTE-increase
    plots to detect moments of relative high inaccuracies and highlight interesting challenges of the dataset.
    Plot A corresponds to a dataset with flickering lights
    and thus the illumination from frame to frame changes drastically. We
    highlight how ORB-SLAM3 requires an abrupt correction due to this. Plot B
    corresponds to a dataset in which the operator is seated and has screens
    with moving content in front of them. Here we can see how the well-performing systems
    OKVIS2 and Basalt, have the worst estimates. Plot C shows how an overexposed
    section breaks SnakeSLAM midway. Plot D features a section with hand occlusions. In plot E we can observe how a
    section with low texture and another with motion blur affect all system
    estimates. Lastly, plot F shows how sections of the datasets with a moving
    person close to the camera manages to produce either very strong jitter or
    diverge the systems. Notice that systems not shown have either diverged or crashed
    before the specific section.}
  \label{fig:trajectories}
\end{figure*}

MSD features challenging scenarios designed to stress test commonly
unaddressed weaknesses in visual-inertial tracking. Handling these cases is
essential when operating with head-mounted XR devices and highly dynamic
humanoid robots. This release includes 64 unique sequences accounting for 5 hours
and 15 minutes of non-calibration footage. Each device dataset contains the
\emph{``Calibration''} and \emph{``Others''} common subcategories. The
calibration procedure is described in further detail in \ref{sec:calibration}.
The Index has a \emph{``Playing''} subcategory with footage taken during
high-intensity gameplay sessions. Given the large amount of sequences, we
provide an overview of their hierarchy with their respective
identifiers and examples in Table \ref{tbl:sequences}. We also contextualize
some notable recordings in this section and present example diagrams of the trajectories in Fig. \ref{fig:trajectories}.
\begin{table}
  \centering
  \caption{Dataset hierarchy and examples}
  \includegraphics[width=0.475 \textwidth]{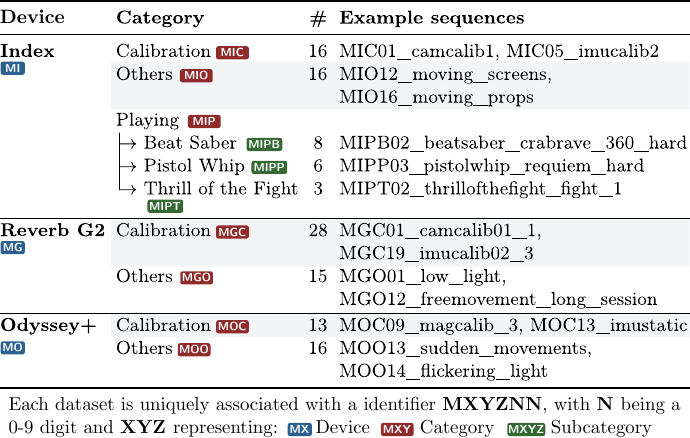}
  \label{tbl:sequences}
\end{table}

The \textbf{Valve Index} \emph{Calibration} subcategory has 8 pairs of
sequences meant for performing camera and IMU-camera calibration as
described in the next section. The subcategory \emph{Others} contains
various footage with different stress-tests like moving persons, multiple
flashing screens, and a changing environment. Lastly, the \emph{Playing} subcategory has
recordings taken while the operator plays three different high-intensity games:
Beat Saber,
Pistol Whip,
and Thrill of the Fight.
Besides the fast-paced and
erratic nature of the trajectories, an additional challenge is the recurring
appearance of hand occlusions (e.g., Fig. \ref{fig:cameras}\,A). A particularly notable sequence here is
\smalltt{MIPB08\_beatsaber\_long\_session\_1}, which has \approxtilde37 minutes of continuous footage and can be used for
assessing tracking drift over extended intensive usage. Such extensive datasets
with continuous high-intensity movement are rare in the literature.%

For the \textbf{HP Reverb G2}, the \emph{Calibration} subcategory comprises 4
sets of pairwise camera, camera-IMU, and magnetometer calibration sequences. The \emph{Other} recordings include,
besides similar ones to the Valve Index, a variety of challenging scenarios.
Some notable examples are \smalltt{MGO01\_low\_light} with low light conditions
increasing motion blur and pixel noise. \smalltt{MGO13\_sudden\_movements}
having very abrupt and sudden head rotations that manage to saturate IMU readings.
\smalltt{MGO14\_flickering\_light} with flickering illumination to test the
robustness of feature trackers. And \smalltt{MGO15\_seated\_screen} targeting
cabin simulation usage with the operator sitting in front of screens with moving
content (Fig. \ref{fig:cameras}\,B). Fig. \ref{fig:trajectories}\,A, B and E
show similar examples of these scenarios for a different device.

Lastly, the \textbf{Samsung Odyssey+} \emph{Calibration} and
\emph{Other} sequences are similar to the previous two devices. An
IMU-only dataset \smalltt{MOC13\_imustatic} is recorded with the headset staying
still for \approxtilde48 hours. This data is meant for performing Allan variance
\cite{allanStatisticsAtomicFrequency1966} analysis of the noise parameters of the IMU. Accurately computing these
can provide an improvement in tracking quality, as shown
in \cite{zhang100PhonesLargeVISLAM2024}. A stationary dataset \smalltt{MOO16\_still} with full sensor data is
also provided for analyzing tracking jitter and drift when no movement
is present in the system.

\subsection{Calibration}
\label{sec:calibration}

We publish both raw and calibrated data. For convenience, each headset presents a
main calibration file, and all their sequences contain pre-calibrated IMU,
ground-truth, and camera timestamps. At the same time, we provide raw data for
all sequences, USB arrival times, alternative calibrations, and several
calibration sequences for each device. Providing raw data is important for
reproducibility and future improvements like the ones found in KITTI
\cite{cvisicRecalibratingKITTIDataset2021}
or EuRoC
\cite{ruckertSnakeSLAMEfficientGlobal2021,mur-artalVisualInertialMonocularSLAM2017}.
Unprocessed data also supports
the development of online methods that can adapt and compensate for the impossibility of
preprocessing the input data in real-time systems.

We devise a calibration pipeline that builds on top of the calibration tools
from Basalt \cite{usenkoBasaltVisualInertialMapping2020,schubertBasaltTUMVI2018}. First, we perform stereo-camera calibration to
estimate the relative pose between each camera and their intrinsics. We provide
estimates for two common distortion models: equidistant
\cite{kannalaGenericCameraModel2006} and radial-tangential
\cite{brownDecenteringDistortionLenses1966,opencv_library}.

For camera-IMU calibration, we use similar sequences with more dynamic movements
that excite all six axes of the IMU. Starting from the previous stereo-camera
calibrations, we optimize the IMU position relative to the cameras,
misalignment, and scale matrices as in \cite{trawnyIndirectKalmanFilter}, and
initial bias estimates for both the gyroscope and accelerometer. This generates
the sensor poses that can be seen in Fig. \ref{fig:schematics}.

We estimate any time offsets that might exist between the sensor clocks by
aligning the rotational velocities of a given trajectory with the gyroscope
measurements of the IMU as explained in \cite{schubertBasaltTUMVI2018}. We use
the ground-truth trajectory for aligning the Lighthouse and IMU clocks, and a
visual-only odometry estimate for aligning the camera and IMU clocks. The Valve
Index benefits the most from this since its cameras and IMU sample timestamps
are not in the same hardware clock. We manually verify that the gyroscope and
the rotational velocity peaks align properly.

Finally, we correct the IMU measurements of all sequences by applying
the estimated misalignment and scale matrices, as well as their initial
bias estimates. We transform the reported ground-truth pose to coincide
with the IMU pose and then apply the timestamp offsets to match the IMU
clock domain in both ground-truth and camera recordings. We perform
multiple sanity checks of the raw and post-processed datasets to avoid
malformed data. The documentation and complementary tools used for this and other
automation tasks needed to replicate this work are publicly released\textsuperscript{\ref{foot:xrtslam}}.

\section{EVALUATION}
\label{sec:evaluations}

We consider multiple systems to evaluate. From this, we initially discard
OpenVINS \cite{genevaOpenVINSResearchPlatform2020}, HybVIO
\cite{seiskariHybVIOPushingLimits2022}, and DPVO \cite{teedDeepPatchVisual2024}.
These systems have shown state-of-the-art performance in their respective
domains, with the first two being filter-based and DPVO learning-based. However,
when running them on a subset of easy sequences from MSD, they either fail to
run them or fail to run them with reasonable accuracy. We do not attribute this
to their underlying methods, but rather to the lack of reference datasets that
cover these kinds of scenarios. Five representative systems are selected and
configured to the best of our knowledge to meet real-time requirements similar
to the ones found in XR or robotics applications. We perform modifications
whenever necessary to obtain \textit{causal} pose estimates
\cite{leuteneggerOKVIS2RealtimeScalable2022}. All modifications and results are
available in the dataset's documentation.

\begin{table}[h!]
  \caption{Systems evaluation}
  \includegraphics[width=0.475\textwidth]{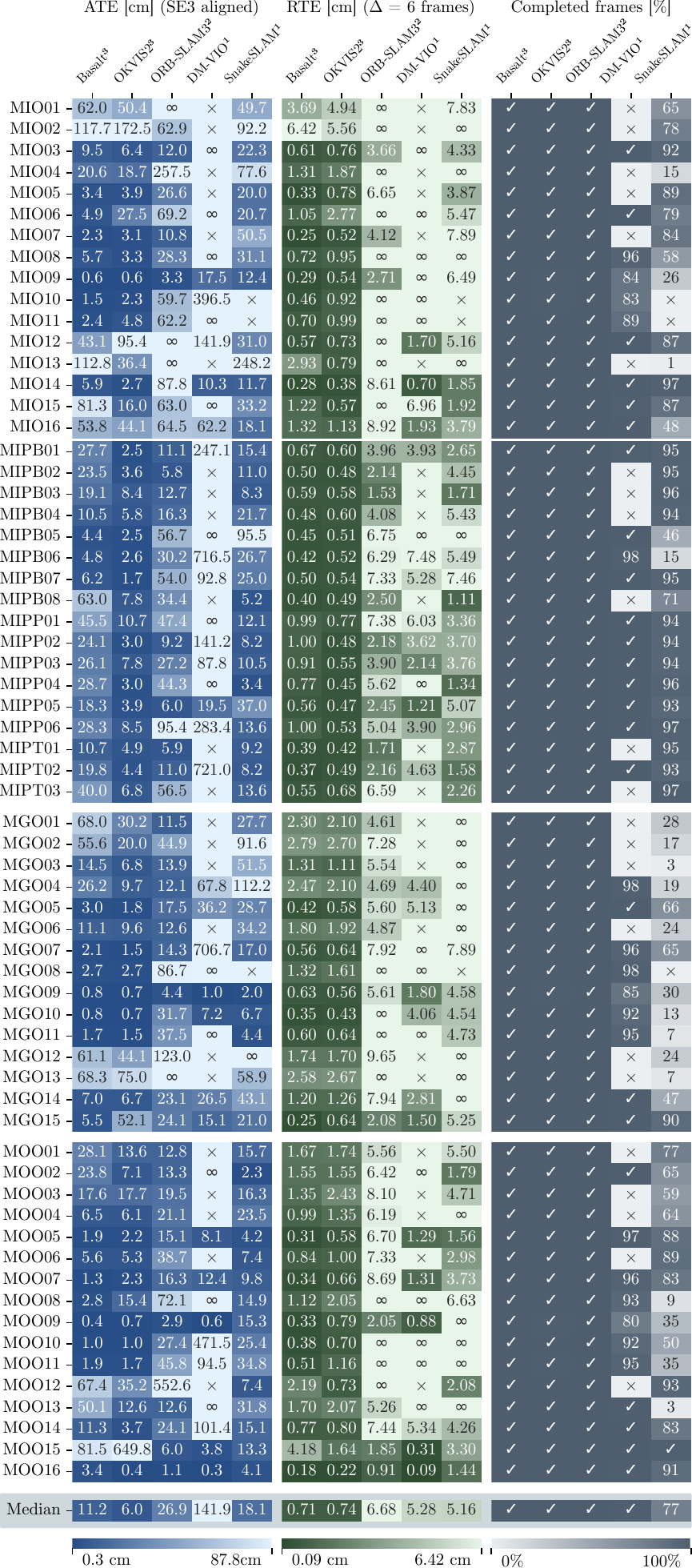}
  \textit{\texttt{×} indicates the system did not produce any estimate.\newline
  \texttt{$\infty$} indicates too big of an error (greater than 10 m ATE or 10 cm RTE).\newline
  \texttt{\checkmark} indicates more than 98\%.\newline
  Sensors:  \;\;\texttt{$^1$}Monocular-IMU  \;\;\texttt{$^2$}Stereo-IMU  \;\;\texttt{$^3$}IMU with 2 or 4 cameras.}
  \label{tbl:benchmark}
\end{table}

\textbf{Basalt} \cite{usenkoBasaltVisualInertialMapping2020} is a semi-direct
VIO system. We use a modified
version\footnote{\url{https://gitlab.freedesktop.org/mateosss/basalt}} with
support for more than two cameras. \textbf{OKVIS2}
\cite{leuteneggerOKVIS2RealtimeScalable2022} is a full SLAM system that also
supports multi-camera datasets, but we run it in its causal VIO-only mode to
reduce its relatively high frame times. \textbf{ORB-SLAM3}
\cite{camposORBSLAM3AccurateOpenSource2021} is a widely established SLAM system
that works with diverse sensor configurations. We adapt it to extract its causal
estimates. \textbf{DM-VIO} \cite{stumbergDMVIODelayedMarginalization2022} is a
monocular VIO system that shows great results on standard datasets like EuRoC.
Finally, \textbf{SnakeSLAM} \cite{ruckertSnakeSLAMEfficientGlobal2021} is a
top-performing system that also shows promising scores on standard datasets. It
does not support non-parallel cameras and is non-causal. We run it in
monocular-IMU mode instead and extract causal estimates.

We perform an extensive evaluation of these five systems on the Monado SLAM
dataset. Table \ref{tbl:benchmark} shows the resulting \textit{absolute
trajectory error (ATE)} and \textit{relative trajectory error (RTE)} metrics
\cite{sturmBenchmarkEvaluationRGBD2012}, together with a percentage of
successfully estimated frames. Given ground-truth poses $R_{i}$ and

Umeyama-aligned estimate
poses $P_{i}$ \cite{umeyamaLeastsquaresEstimationTransformation1991}, both in $SE(3)$, with $i = 1, \ldots, N$ being the frame index and $N$ the number of
frames in a sequence. The ATE for the sequence is defined as

\begin{equation}
  E_{ATE} = \sqrt{\frac{1}{N} \sum_{i=1}^{N} \left\Vert trans(P_i^{-1} R_i) \right\Vert^2}.
\end{equation}

Where $trans(\cdot)$ is the translation part of the pose. RTE is

\begin{align}
  E_{RTE} &= \sqrt{\frac{1}{M} \sum_{j=1}^{M-1} \left\Vert trans(\delta(P, j)^{-1} \; \delta(R, j)) \right\Vert^2}, \\
  \delta(X, k) &= X_{k\Delta}^{-1} X_{(k+1)\Delta},
\end{align}

Where $\Delta$ is a small frame interval in which the errors are measured and
$M = \lfloor N / \Delta \rfloor$. We use $\Delta = 6$ frames.

From the median ATE estimates we can see that OKVIS2 shows the lowest error,
followed by Basalt. In particular, OKVIS2 outperforms Basalt in the \texttt{MIP}
gameplay sequences. The other systems have larger errors and even tracking failures.
Notice from the medians, that more than 50\% of all ATE errors are greater than 5
cm, a value that can be considered too high for the XR use case and for applying
fine motor skills in humanoid robots. It is worth noting that, if allowed to run
non-causally, OKVIS2, ORB-SLAM3, and SnakeSLAM show much better median ATE
scores of 2.2 cm, 2.5 cm, and 2.3 cm, respectively. However, this is not a
reasonable operation mode for real-time applications.

The RTE metric is particularly important for XR since it impacts how well
tracking estimates are perceived by a user wearing a headset. For robotics,
it determines the limits of accuracy during fast robot movements. In this
metric, Basalt and OKVIS2 show the lowest errors again. Even then, we
can observe errors greater than 1 cm in many sequences, which can be too high for our target
applications. The errors from the other systems
are an order of magnitude larger. Finally, the completed frames column shows the
percentage of frames that were successfully estimated by each system. Basalt,
OKVIS2, and ORB-SLAM3 estimate all frames, however it is worth
noting that ORB-SLAM3 crashes on 2\% of the underlying runs. DM-VIO fails in
many, while SnakeSLAM provides very sparse valid estimates with some failures.

It is worth noting the effect of individual recordings in each system. Sequences
like \smalltt{MIO13\_moving\_person} (similar to \ref{fig:trajectories}\,F) or \smalltt{MGO13\_sudden\_movements} (similar to Fig. \ref{fig:trajectories}\,E) seem to
be an unsolved challenge for all systems. Others like
\smalltt{MOO15\_seated\_screen} show the advantages of methods like DM-VIO
that underperform in other sequences as shown in Fig. \ref{fig:trajectories}\,B.

\begin{figure}
  \centering
  \includegraphics[width=0.475\textwidth]{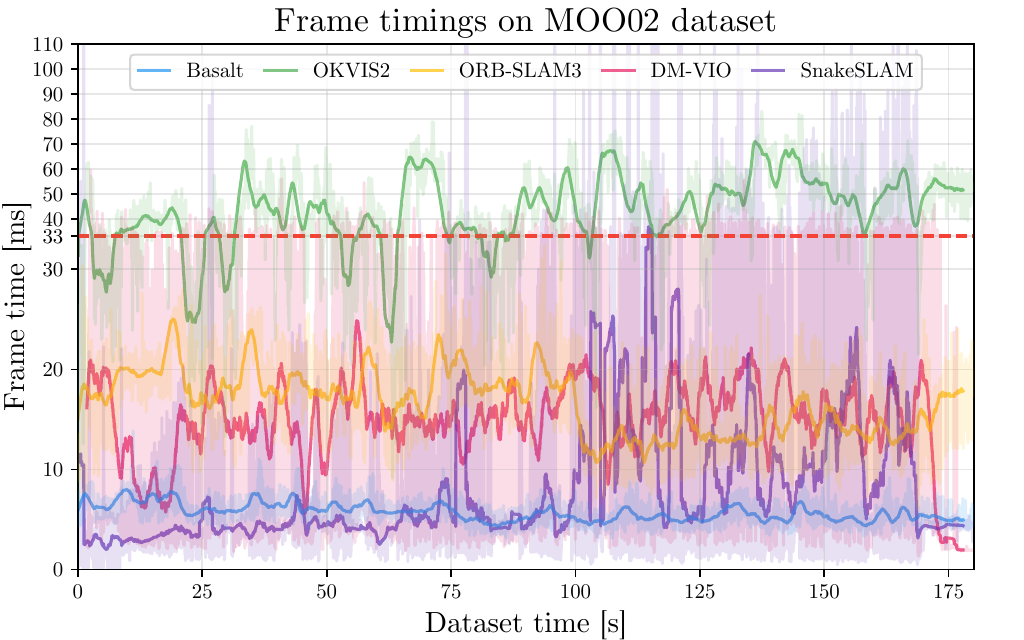}
  \caption{Time spent per stereo frame in the \smalltt{MOO02\_hand\_puncher\_2}
  dataset on an Intel Cora Ultra 7 258V laptop processor. We show a moving average
  with a window size of 2 seconds (60 frames). In the background, the raw data
  shows outliers and jitter for each system. A red line marks the time limit for
  real-time operation for the Odyssey+ headset. Note that DM-VIO and SnakeSLAM
  run in monocular-IMU mode instead of stereo-IMU.}
  \label{fig:timings}
\end{figure}

For the timing analysis, we consider the \smalltt{MOO02} dataset. Fig.
\ref{fig:timings} shows the computation time each system spent for every frame.
Basalt seems to take the lead in this analysis. OKVIS2 is the only
system having trouble keeping its computation time under the 33 ms needed for
real-time operation on the Odyssey+. ORB-SLAM3 and DM-VIO perform
similarly, although DM-VIO is processing only one camera. SnakeSLAM performance
is the best at the beginning, but it deteriorates as the sequence progresses, with
many outliers that go above 33 ms.

Besides divergences and crashes, ATE values greater than 10 cm are observed across all
systems in more than 50\% of the total runs. Avoiding such inaccuracies remains
difficult for current systems. This is also reflected in the RTE, timing, and completion
metrics, where most systems struggled to provide reliable performance across all
runs. These results show the need for further advancements to improve accuracy,
efficiency, and robustness for real-time applications.

\section{CONCLUSION}
In this work, we present the Monado SLAM dataset, a novel visual-inertial
dataset covering emerging use-cases like XR and humanoid robotics. It contains
challenging sequences that are designed to highlight the limitations of current
state-of-the-art systems. Public datasets from XR devices are scarce, with only
two AR devices available in the literature, making MSD the first publicly
available VR dataset. We devise a cost-efficient procedure for dataset creation
leveraging commercial XR devices with sequences from three distinct headsets.
More than 5 hours and 15 minutes of samples are recorded with dense ground-truth
from external sensors. We publish our dataset under a permissive CC BY 4.0
license to foster collaborations in an academic field closely connected to industry
applications.

We additionally present a thorough evaluation of diverse systems. Besides ATE,
we report commonly overlooked metrics that are crucial for real-time operation.
The results of the benchmark show that there is still many open challenges to
address both accuracy and efficiency problems for real-world robotics and XR
applications. In this way, MSD stands as a valuable resource for the research of new
and existing visual-inertial tracking methods.

\section{ACKNOWLEDGEMENTS}

We express our appreciation to the Monado maintainers and its community. We
thank Collabora Ltd. for funding the recording of the dataset. This work was
supported by the ERC Advanced Grant SIMULACRON and by the Munich Center for
Machine Learning.

\balance

\end{document}